\pdfoutput=1

\documentclass[11pt]{article}

\usepackage{EMNLP2023}
\usepackage{graphicx}
\usepackage{caption}
\usepackage{placeins}
\usepackage{float}
\usepackage{times}
\usepackage{latexsym}
\usepackage{booktabs}
\usepackage{subcaption}
\usepackage{multirow}
\usepackage{multicol}
\usepackage{amsmath}

\usepackage[T1]{fontenc}

\usepackage[utf8]{inputenc}
\usepackage{verbatim}
\usepackage{microtype}

\usepackage{inconsolata}

\title{Audio-Based Crowd-Sourced Evaluation of Machine Translation Quality}

 \author{Sami Ul Haq$^{1,2}$, Sheila Castilho$^{1,2}$, Yvette Graham$^{2,3}$\\
         $^1$ADAPT Centre\\
         $^2$Dublin City University (DCU), Ireland\\
         $^3$Trinity College Dublin (TCD), Ireland\\
         {\tt \{sami.haq, sheila.castilho, yvette.graham\}@adaptcentre.ie}
         }

\begin{document}
\maketitle
\begin{abstract}
Machine Translation (MT) has achieved remarkable performance, with growing interest in speech translation and multimodal approaches. However, despite these advancements, MT quality assessment remains largely text-centric, typically relying on human experts who read and compare texts. Since many real-world MT applications (e.g., Google Translate Voice Mode, iFLYTEK Translator) involve translation being spoken rather printed or read, a more natural way to assess translation quality would be through speech as opposed text-only evaluations. This study compares text-only and audio-based evaluations of 10 MT systems from the WMT General MT Shared Task, using crowd-sourced judgments collected via Amazon Mechanical Turk. We additionally, performed statistical significance testing and self-replication experiments to test reliability and consistency of audio-based approach. Crowd-sourced assessments based on audio yield rankings largely consistent with text-only evaluations but, in some cases, identify significant differences between translation systems. We attribute this to speech’s richer, more natural modality and propose incorporating speech-based assessments into future MT evaluation frameworks.

\end{abstract}

\section{Introduction}
Reliable evaluation process is critical in the development and refinement of MT systems. MT evaluation (MTE) often relies on both automated and manual measurement techniques. Manual evaluation is always a preferred choice and provides a deeper understanding of system quality, while automatic evaluation metrics (AEMs) often serve as a proxy for human judgments \citep{CASTILHO_2018_TQA}. AEMs support reusable assessments, system comparison and rapid MT deployment. However, AEMs face several issues including their inability to handle contextual and cultural nuances, the dependency on reference translation, and domain-specific challenges. Therefore, despite being time-consuming and expensive, human assessment is still a fundamental requirement for reliable evaluation. 

The annual Conference on Machine Translation (WMT) is the primary forum for collecting human judgments to evaluate metrics and participating systems in its shared translation task each year. In early evaluation campaigns, 5-point adequacy and fluency ratings were gathered from participants as the primary evaluation metric \citep{koehn2006manual}. Subsequent WMT campaigns adopted a ranking-based evaluation approach as the official metric \citep{vilar2007human}, with rankings still collected from participants of the evaluation campaign. Regarding fluency as a measure of MT output quality, \citet{graham_continuous_2013} argued that using a 1-100 continuous scale yields better inter-annotator consistency compared to a five-point interval scale. Supporting this, \citet{bojar2016findings} found strong correlations between adequacy and fluency-based evaluations. These findings led WMT to replace relative ranking with adequacy-based Direct Assessment (DA) on a continuous scale as the official metric \citep{bojar2017findings}. For into-English translation tasks, WMT frequently relied on crowd-workers for its human evaluation campaigns. Crowd-based evaluations allow for a fast and cheap MT quality evaluations \citep{callison2009fast}. When coupled with quality-controlled annotations, non-expert crowd assessments show better inter-annotator consistency \citep{graham_continuous_2013, graham_can_2017}. However, \citet{castilho2017crowdsourcing} found that crowd-workers, compared to professional translators, were less capable of detecting subtle MT errors. Studies by \citet{laubli-etal-2018-machine} and \citet{toral_attaining_2018} also favored the use of professional translators over researchers or crowd-workers due to their ability to differentiate between human and machine translations. Consequently, WMT revised its evaluation procedures to prioritize professional translators over crowd-workers \citep{kocmi_findings_2022, kocmi_findings_2023}. Despite its limitations, crowd-based assessment remains the most convenient choice for certain tasks, particularly monolingual DA, which does not require human raters to have bilingual knowledge \citep{graham_can_2017}, making it easier to conduct. More recently, WMT performed evaluations using Error Span Annotation (ESA) protocol \citep{kocmi2024error}, which requires annotators to assign an overall score to each segment, similar to DA and classify errors based on severity (e.g. major or minor).  

The human evaluation process has evolved over time; however, there is still no consensus on the best approach to evaluating translation quality \citep{CASTILHO_2018_TQA}. Current MT evaluation metrics primarily considers text, despite the fact that many real-world MT applications involve spoken rather than written translation. Most importantly, the recent emergence of pre-trained multimodal models \citep{barrault2023seamlessm4t} has enabled support for direct speech-to-speech, text-to-speech and speech-to-text translation, however appropriate methods for evaluation for these systems are yet limited or borrowed from text-domain \citep{salesky2021assessing, sperber_evaluating_2024}. 

We argue that speech, as a natural and expressive modality, can provide more reliable measures of MT quality. To support this claim, we propose incorporating TTS technology into direct MT assessment, allowing for a direct comparison between text-only and speech-enabled evaluation approaches. Our study collects human judgments for German-English translations from WMT shared task using crowd-workers hired via Amazon Mechanical Turk. The evaluation consists of two conditions: (i) a text-only setup, replicating the conventional method where evaluators compare written MT output with a reference translation, and (ii) a text-audio setup, where evaluators listen to the MT output while reading the reference translation. We perform self-replication experiments and statistical significance tests to assess the consistency and reliability of the proposed method.

A comparative analysis of these evaluation conditions yields two key findings. First, rankings derived from text-audio evaluations are broadly similar to the original evaluations but also show notable differences compared to conventional setups, with the audio-based method demonstrating a substantially greater ability to detect significant differences between translation systems. We hypothesize that this difference arises because speech is a natural and rich modality, capable of conveying prosodic and expressive features that text alone cannot capture. Second, consistent with prior research, our results confirm that crowd-workers tend to assign lower rankings to human translations that diverge from the reference, while favoring literal machine translations \citep{castilho-etal-2017-comparative, fomicheva_role_2017}. Furthermore, self-replication experiments reveal a higher positive correlation between repeated runs of audio-based evaluations, indicating improved reliability and consistency of this new approach. 

\section{Methodology}

\subsection{Data set}

We used MT outputs from WMT 2022 German-English translation task, comprising around 20,000 translations submitted by 10 participating systems, with each system contributing approximately 2,000 translations. This original evaluation set is a bilingual corpus drawn from different domains, as shown in Table~\ref{corpora-domain}, with document lengths varying considerably by domain. To ensure balanced domain representation while preserving document order, a subset of documents was randomly sampled from each domain. We use on average 450 segments per system for multimodal\footnote{In this study, multimodal is used to refer to text-audio based setup} and text-only experiments.

The WMT evaluation campaign has already published results from crowd-based human evaluations of the submitted systems. As WMT now conducts bilingual ('source-based') evaluations using professional translators, we focus on WMT 2022—the most recent workshop to perform monolingual DAs.

\begin{table}
\centering
\begin{tabular}{lcr}
\hline

\textbf{Domain} & \textbf{\#segments} & \textbf{Avg. doc length}
\\
\hline

\verb|conversation| & \verb|462| & \verb|6.8| \\
\verb|ecommerce| & \verb|501| & \verb|18.5| \\
\verb|news| & \verb|506| & \verb|14.5| \\
\verb|social| & \verb|515| & \verb|15.6| \\
\hline
\end{tabular}
\caption{\label{corpora-domain} Number of segments and average document length (\#segments per document) of German-English data used in the general translation test sets.}
\end{table}


\subsection{Assessment Design}
AMT crowd-sourcing service was used to design and collect human judgments, with each task consisting of 100 segments.  A single segment along with a reference translation is presented at one time. Where possible, segments are collected and shown in document context. In adequacy based assessments, crowd-workers are asked to rate how adequately an MT output expresses the meaning of reference translation. The scores are collected on 0--100 visual analog scale (VAS) for each segment. Additionally, rater quality control mechanism is implemented to filter out ratings from non-reliable raters, as outlined by \citet{graham_can_2017}. At the end of the task, evaluators have the option to provide feedback on their experience. 

The segment-level ratings were used to calculate system-level rankings. At the end of the evaluation, we provide two types of segment-level scores, averaged across one or more raters: raw scores and z-scores, with the latter standardized for each annotator. The final score of an MT system is the mean standardized score of its ratings after filtration. Multiple judgments are collected per segment, increasing the number of annotators per translation enhances the consistency and reliability of the mean score.
Since reference-based assessment required only knowledge of the English language; the selection criteria required participants to be native English speakers.



\begin{figure*}[t]
  \centering  \fbox{\includegraphics[width=0.9\linewidth]{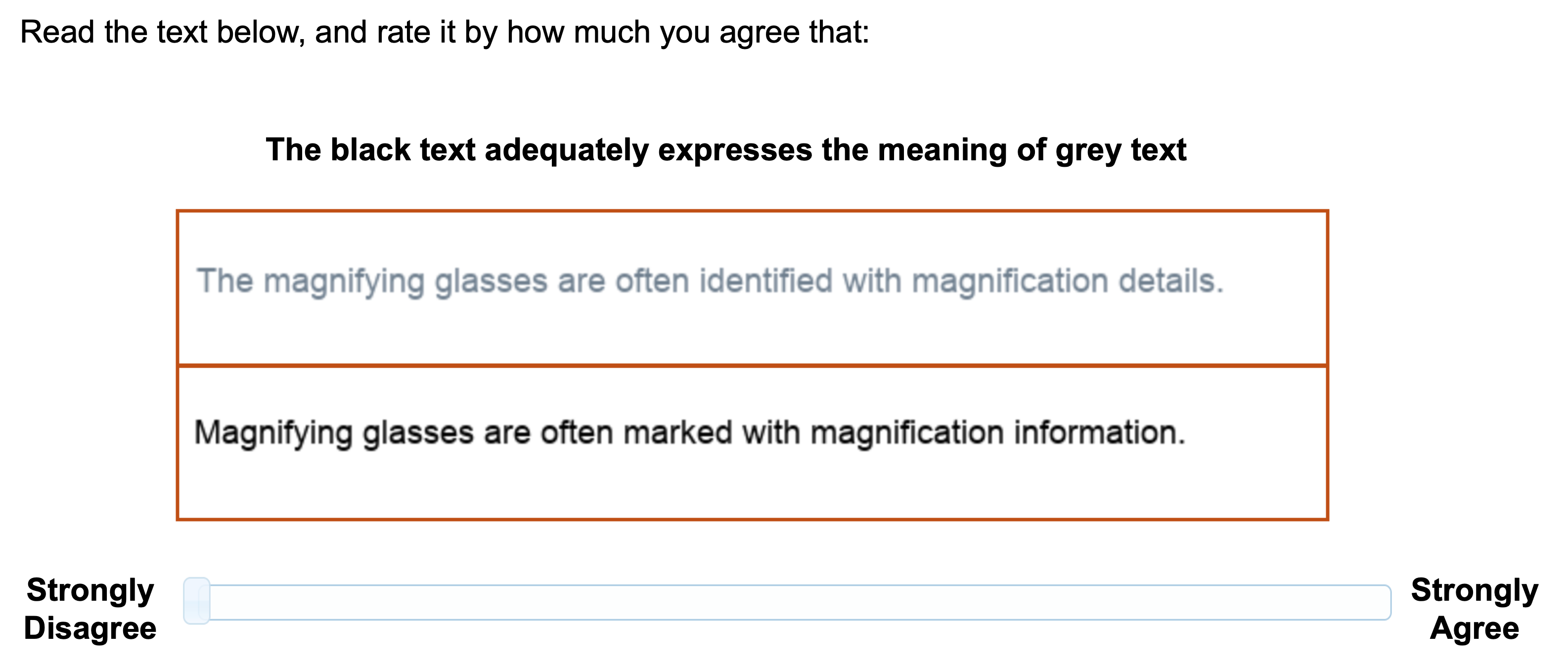}}
  \caption{Screenshot of the text-only assessment interface, as presented to an AMT worker. Reference text is presented in grey while MT output is shown in black text. The slider is initially placed at left most corner; workers move it to the right in reaction to the question.}
  \label{fig:uni-modal_interface}
\end{figure*}

\begin{figure*}[h]
  \centering
\fbox{\includegraphics[width=0.9\linewidth]{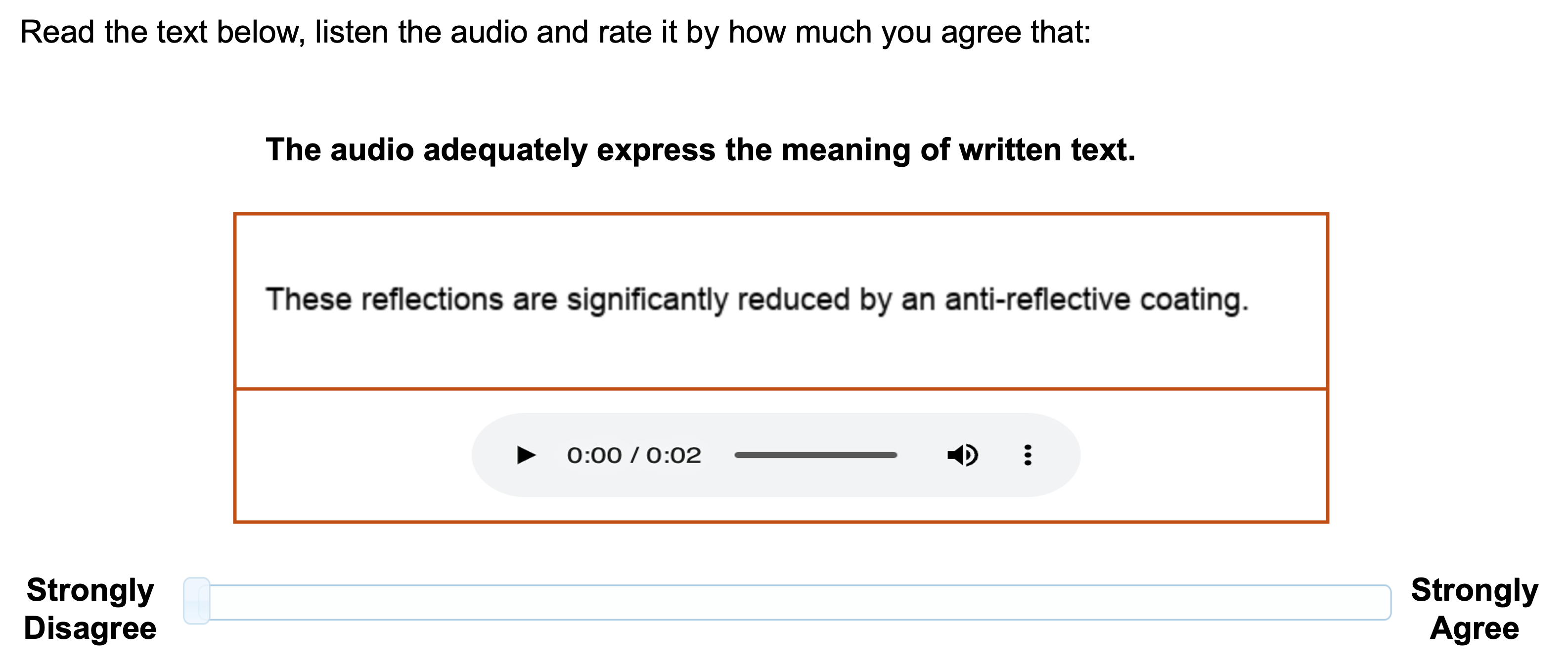}}
  \caption{Screenshot of the multimodal assessment interface, as presented to an AMT worker. Worker can use audio control to listen translations, the text in presented in the image form. The slider is initially placed at left most corner; workers move it to the right in reaction to the question.}
  \label{fig:multi-modal_interface}
\end{figure*}

We compare judgments collected using following two different setups: 
\begin{itemize}
    \item \textit{Text-only}: MT output and reference translations, both are presented as text (Figure \ref{fig:uni-modal_interface}).
    \item \textit{Multimodal}: MT output is presented in audio (TTS) and reference translation as text (Figure \ref{fig:multi-modal_interface}).
\end{itemize}

Overall, we gathered approximately 12,000 crowd-sourced judgments for English-German language pair using DA. Compared to ordinal ranking or relative preference judgments \citep{callison2009fast}, direct estimation facilitates more robust statistical analysis, thus making it suitable for crowd-sourced annotations \cite{graham_continuous_2013}. When combined with quality control mechanisms, direct assessments have shown effective and relatively consistent human judgments of MT quality in WMT evaluation campaigns \citep{specia-etal-2020-findings, akhbardeh_findings_2021, kocmi_findings_2022}. 

\subsubsection{Text-only setup}
We randomly sampled 500 segments per system (with the addition of quality control segments, the total could be increased). The selected translations are then converted into bit-mapped images,  in order to deter workers from using speech feature of Web Browsers to read-aloud the translations.

In this scenario, the workers are shown the reference and the MT output as text and asked to rate MT output by moving the slider (as shown in Figure \ref{fig:uni-modal_interface}). For task simplicity, we kept the structure of assessment similar to existing evaluation setups \citep{graham_can_2017, kocmi_findings_2022}. The ratings are collected per segment in a sequential manner, adhering to the document order where feasible. However, longer documents may need to be divided into smaller units to comply with the limit of 100 segments per task. The setup restricts assessors from revisiting and modifying ratings of previous segments to ensure integrity of quality control measures. 

\subsubsection{Multimodal setup}

For comparison, the same segments sampled for the text-only scenario were considered in this experiment. However, this setup utilises TTS technology to present the MT output in an audio-equivalent form. To make the task less cognitively taxing, we present only the MT system's output in audio form. For this,  we used the Google Cloud Text-to-Speech (TTS) Service (GCS)\footnote{\url{https://cloud.google.com/text-to-speech}} to generate audio representations of MT outputs. The service was employed with its default human-like voice settings, which are noted for their high quality and clarity. 
GCS is well-suited for long-form content\footnote{For multimodal experiments, in total a human assessor may have to listen up-to 20 minutes of machine translation outputs, therefore along with accuracy of TTS, a pleasant listening experience is important.} due to its close approximation of human speech and its ability to provide an enhanced listening experience \cite{cambre_choice_2020}.

\subsubsection{HITs}
Both multimodal and text only assessments are carried out separately. Each task, referred to as “HITs” (Human Intelligence Task) contains 100 translations in total for each setup. In addition to system output, a set of quality control segments was added, keeping the total size of HIT to 100. The quality control segments consists of exact repeats (\textit{ask\_again}) and degraded translations (\textit{bad\_reference}), duplicated from system outputs. Thus, each HIT consists of approximately 20\% quality control segments (used to estimate workers' reliability) and 80\% genuine system outputs. To create \textit{bad\_reference} pairs, we followed the strategy of randomly substituting words in a sentence, as outlined in \citet{graham_crowd-sourcing_2013}. For the multimodal setup, the quality control segments were first prepared using the same strategy in text form and then converted into audio using the TTS API.

Judgments from crowd workers with limited or no knowledge of the assigned task pose a significant risk of inconsistency and discrepancies in the results. Expert-based MT quality assessment is the preferred approach; however, it incurs high economic and time costs, making crowd-sourcing a viable alternative. Consequently, assessing worker reliability becomes critically important in crowd-sourced evaluations. Quality control segments within HIT allow for reliability estimates based on workers distribution of scores assigned to \textit{bad\_reference} and \textit{ask\_again} items. These estimates are based on following two assumptions: 
\begin{enumerate}
    \item The consistent assessor will assign significantly higher score to the system producing high quality translations compared to a system producing inferior outputs. 
    \item The consistent assessor will assign highly similar scores in repeated evaluations of the same translations.
\end{enumerate}

\begin{figure}[h!]
    \centering
    \includegraphics[width=1\linewidth]{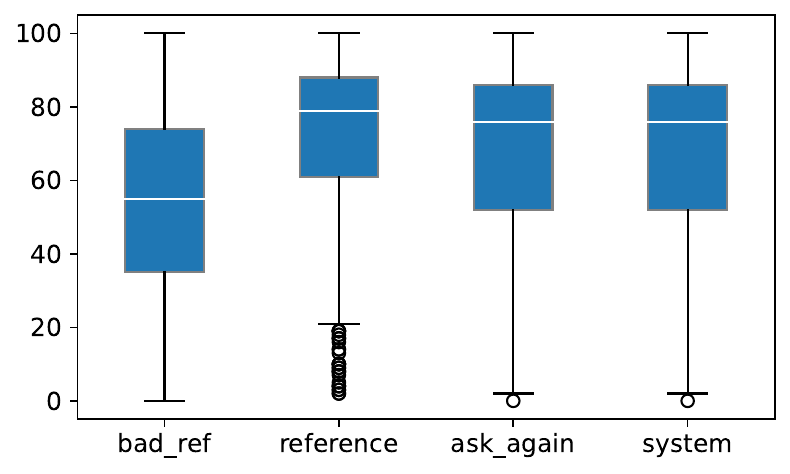}
    \caption{ Score distributions of reliable workers across different quality control segments: $bad\_reference$, $ask\_again$, and original system outputs.}
    \label{fig:good-worker-stats}
\end{figure}

Analysis of assumptions 1 and 2 can provide a measure of workers' ability to differentiate between a good and inferior translation. Assumptions 1 and 2, based on the sets of \textit{bad\_reference} and \textit{ask\_again} translations, posit that a consistent worker would assign significantly lower scores to degraded (\textit{bad\_reference}) translations and similar scores to repeated (\textit{ask\_again}) translations. For this, we apply the Wilcoxon rank-sum test to compare the score differences between \textit{ask\_again} and \textit{bad\_reference} translation pairs, with a resulting $p$ value as an estimate of reliability. The expectation is that the difference in scores for degraded translation pairs will be smaller than for repeated judgments. A lower p-value ($p$ < 0.05) indicates higher reliability, demonstrating that the worker can effectively distinguish between high-quality and degraded translations. As shown in Figure \ref{fig:good-worker-stats}, conscientious workers assigned lower scores to degraded translations compared to the original references. Furthermore, for repeated segments, they exhibited a consistent scoring pattern by assigning similar scores to identical pairs.

Table \ref{tab:quality_control_worker} provides statistics on the number of workers involved in each assessment type and the percentage of workers who passed the quality control threshold. A similar trend was observed across both assessment types, with nearly 20\% of workers meeting the reliability criteria. To determine whether to accept or reject HITs, the mean score differences for \textit{bad\_reference} and \textit{ask\_again} pairs were carefully analyzed, rather than relying solely on automatic quality control checks.\footnote{In addition to statistical tests, other measures were in place to detect robotic or low-quality submissions, such as extremely short completion times, lack of slider movement, and assigning the same rating to every judgment.} This is further reflected in the difference between the number of approved workers and those who met the quality control criteria. Rejected HITs were rescheduled to obtain fresh judgments.

\begin{table*}
   \centering
   \begin{tabular}{lcccccc}
      \toprule
      & \multicolumn{3}{c}{Workers} & \multicolumn{3}{c}{Translations} \\
      \cmidrule(lr){2-4} \cmidrule(lr){5-7}
      modality   & Total & Approved & Pass QC & Total & Approved & Pass QC  \\
      \midrule

     text only  & 225 & 47  & 42 (18.50\%) & 23.3k & 5.1k  & 4.6k (19.7\%) \\

      multimodal  & 242 & 52  & 48 (19.83\%) & 26.1k & 6.0k & 5.3k (20.3\%)\\

      \bottomrule
   \end{tabular}
   \caption{Numbers of workers and translations, before and after quality control for multimodal and text only experiments.}
   \label{tab:quality_control_worker}
\end{table*}

\section{Results} 
System rankings are calculated for each setup using filtered judgments—only those that passed the quality control criteria. Quality control segments (\textit{bad\_reference} and \textit{ask\_again}) are excluded from the final system rankings. System rankings are based on the mean raw and standardized ($z$) scores. To compute the standardized score for each system, individual scores are first normalized using each worker’s mean and standard deviation (as per equation \ref{eq:z-score}). The standardized scores for all segments corresponding to a system are then averaged to obtain the system-level score \citep{graham2014randomized}. Since HITs are structured so that a single worker may assess multiple systems, standardising the scores helps mitigate individual biases and harmonise outputs across workers.

Table \ref{ranking_all} presents the raw and standardized scores of the participating systems across different experiments, with the last three columns showing the official results from WMT. Systems are ordered from best to worst based on their average standardized scores, with the raw score used as a secondary criterion when standardized scores are identical to two decimal places.
\begin{equation}
    z = \frac{x - \mu}{\sigma}
    \label{eq:z-score}
\end{equation}

The text-only results show increased correlation between the raw and standardised score, with few exceptions such as system \texttt{LT22} and \texttt{JDExploreAcademy}, which would have ranked better according to raw score. This close correlation suggests an even distribution of segments from different systems across workers and may also be attributed to the homogeneous nature of the task (text-only data). It is important to note that these rankings may not fully reflect actual system performance or align with the official WMT rankings, as we used a smaller set of judgments per system compared to WMT22, with the primary objective of investigating and comparing audio-based and text-based evaluations.

In the multimodal scenario, the differences in system rankings between $z$ scores and raw scores are more pronounced. Based on the raw scores, a different ranking emerges, with \texttt{Lan-Bridge} performing best. This divergence may be caused by the differing nature of the evaluation setup, particularly the use of both audio and text for evaluation. The out-of-sequence numbers in order column (Table \ref{ranking_all}) highlight differences in system rankings across different experiments. 

\begin{table*}[htbp]
\centering

\begin{tabular}{l|ccc|ccc|ccc}
\multirow{2}{*}{System} & \multicolumn{3}{c|}{text} & \multicolumn{3}{c|}{official-text} & \multicolumn{3}{c}{multimodal} \\
\cline{2-10}
 & raw ave. & ave. $z$ & order & raw ave. & ave. $z$ & order & raw ave. & ave. $z$ & order \\
\hline
PROMT & 73.05 & 0.14 & 3 & 66.02 & -0.127 & 10 & 69.63 & 0.19 & 1 \\
Online-G & 68.81 & 0.06 & 6 & 64.1 & -0.057 & 4 & 68.76 & 0.19 & 2 \\
Online-A & 74.06 & 0.19 & 2 & 67.3 & -0.070 & 5 & 74.61 & 0.18 & 3 \\
Online-W & 74.16 & 0.22 & 1 & 70.8 & -0.023 & 2 & 70.74 & 0.14 & 4 \\
Online-Y & 72.16 & 0.13 & 5 & 66.5 & -0.089 & 7 & 69.89 & 0.14 & 5 \\
Online-B & 71.49 & 0.14 & 4 & 66.3 & -0.092 & 8 & 67.10 & 0.08 & 6 \\
JDExploreAcademy & 72.89 & 0.05 & 8 & 68.1 & -0.038 & 3 & 67.85 & 0.07 & 7 \\
LT22 & 74.36 & 0.05 & 7 & 64.8 & -0.126 & 9 & 64.52 & 0.07 & 8 \\
Lan-Bridge & 68.29 & 0.05 & 9 & 68.8 & 0.004 & 1 & 71.24 & 0.04 & 9 \\
Human-B & 66.94 & -0.12 & 10 & 68.3 & -0.086 & 6 & 63.45 & -0.16 & 10 \\
\end{tabular}

\caption{Comparison and system rankings based on scores from the text-only and multimodal (text + audio) setup for the German–English translation direction. Systems are ordered by their average standardized (\textit{z}) scores. In cases of a tie in \textit{z} scores, the average raw (raw ave.) score is used as a secondary ranking criterion.}
\label{ranking_all}
\end{table*}


A direct comparison of the mean standardized scores across both tables reveals substantial differences in system rankings. For example, in the text-only evaluation, \texttt{Online-W} outperforms \texttt{PROMT} based on standardized scores, whereas the multimodal evaluation ranks \texttt{PROMT} as the top-performing system. Similarly, \texttt{Online-G} is ranked sixth, below \texttt{Online-A}, \texttt{Online-W}, and \texttt{Online-Y} in the text-only setup, but is rated higher than these systems in the multimodal evaluation. For most other systems, rankings diverge by one or two places between setups, with the exception of \texttt{Human-B}, which consistently ranks as the lowest-performing system in both evaluations. Ideally, \texttt{Human-B} (the human reference translation) should be the top-performing system. However, the results suggest that crowd-workers struggled to distinguish between human translations and MT outputs. This aligns with prior research suggesting that crowd-workers tend to favor literal, straightforward translations, resulting in lower rankings for human translations that deviate from the reference \citep{fomicheva_role_2017, freitag_experts_2021}.

\begin{figure*}[t]
    \centering \includegraphics[width=0.8\linewidth]{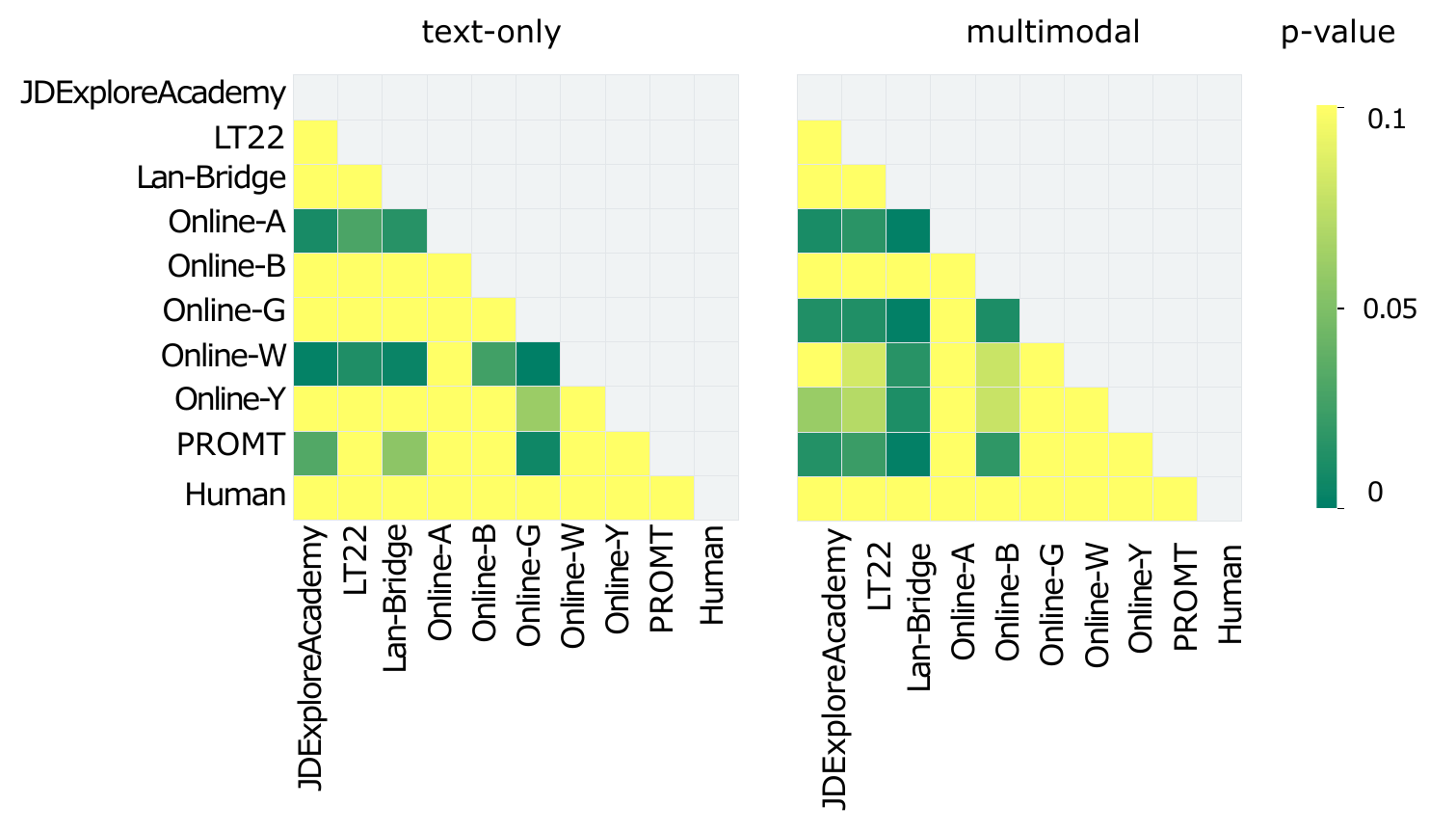}
\caption{Significance test outcomes for text-only and multimodal method of human evaluation. Colored cells indicate that the scores of the row $i$ system are significantly greater than those of the column $j$ system.}
    \label{fig:sig-heatmap}
\end{figure*}

\subsection{Significance Test Results}
Since both approaches yield different system rankings without a clear indication of which better reflects actual performance, more robust testing is required to determine whether the observed ranking differences are genuine. To address this, we employ two techniques: statistical significance testing and self-replication. Significance testing estimates the likelihood that ranking differences between system pairs occurred by chance, while self-replication examines the reproducibility of results to verify their reliability and consistency.

The results of significance tests are visualised as heat maps in Figure \ref{fig:sig-heatmap} for the multimodal and text-only setups. Specifically, we apply one-sided Wilcoxon rank-sum test to compare the standardized human assessment score distributions for each pair of systems.

Tables with head-to-head comparisons between all systems are included in Appendix \ref{appendix_head}.

The significance matrices are constructed under the hypothesis that the scores of system X are significantly better than those of system Y at a given confidence level, $p$. A comparison of the text-only and multimodal heat maps reveals that the multimodal approach results in a slightly higher proportion of significant differences between systems with fewer uncertainties. For example, at a confidence level of $p < 0.05$, the text-only  method identifies relatively few significant differences, whereas the multimodal method  demonstrates more distinct separations among systems. For example, the multimodal heat map shows that \texttt{Online-G} performs significantly better than \texttt{JDExploreAcademy}, \texttt{LT22}, \texttt{Lan-Bridge}, and \texttt{Online-B}, as confirmed by its higher multimodal average \textit{z}-score. Similarly, for \texttt{Online-A}, both the text-only and multimodal evaluations lead to similar conclusions. 

\begin{figure}[h]
    \centering
    \begin{subfigure}[b]{\linewidth}  
        \centering
        \includegraphics[width=0.9\linewidth]{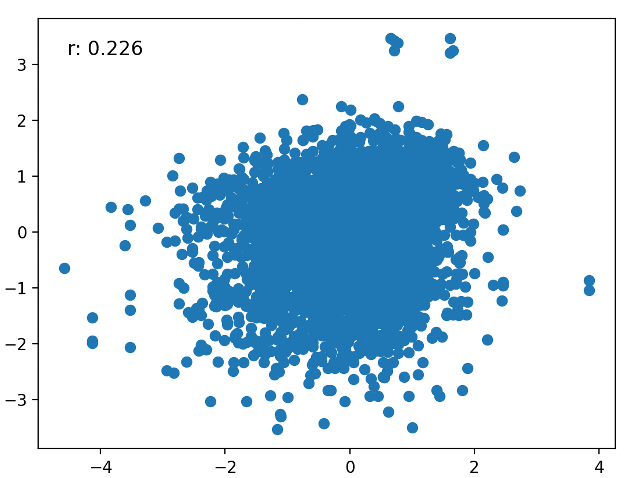}
        \captionsetup{labelformat=empty} 
        \caption{ (a) Self-replication (multimodal vs multimodal)}
    \end{subfigure}
    
    \begin{subfigure}[b]{\linewidth}
        \centering
        \includegraphics[width=0.9\linewidth]{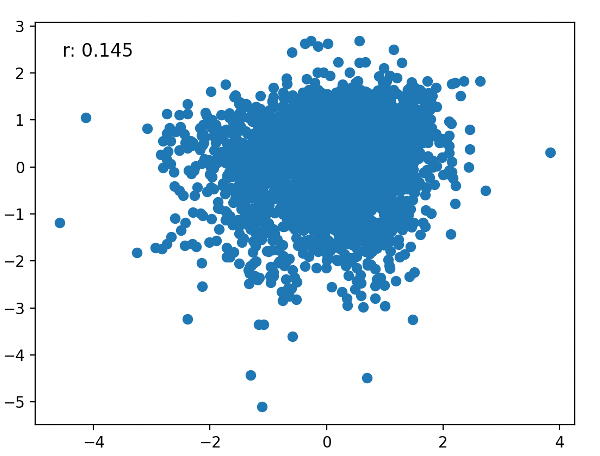}
        \captionsetup{labelformat=empty}
        \caption{(b) text-only vs multimodal}
    \end{subfigure}
    
    \caption{Scatter plots illustrating the correlation ($r$) between different evaluation approaches. (a) shows the correlation between two runs of the multimodal approach, while (b) compares the results of the text-only and multimodal approaches.}

    \label{fig:correlation-scatter}
\end{figure}

\subsection{Self-replication Results}
Figure \ref{fig:correlation-scatter} presents scatter plots comparing initial and self-replicated judgments from multimodal and text-only experiments. To assess the consistency of judgments collected using the multimodal (text and audio) approach, we conduct two independent runs and compute the Pearson correlation ($r$) between the initial and self-replicated results. In Figure \ref{fig:correlation-scatter} (a), self-replicated and original multimodal assessments are plotted on the $x$-axis and $y$-axis, respectively. Figure \ref{fig:correlation-scatter} (b) illustrates the correlation for text-only and multimodal scores, with the former on the $x$-axis and the latter on the $y$-axis. A high correlation would be indicated by points closely aligning with a straight line. While both approaches show a weak positive correlation, the multimodal setup exhibits a slightly higher correlation than the text-only setup, suggesting the potential of audio-based evaluation for providing reliable MT quality estimates.

\subsection{Discussion}
The results of significance and correlation tests suggest that speech can offer consistent and valuable insights into MT quality. We hypothesize that these differences arise because speech is a richer modality, capable of conveying prosodic and expressive features \citep{kraut1992task}. As a result, evaluators listening to translations were better able to detect major variations and unnatural-sounding MT outputs. Furthermore, feedback from evaluators at the end of the assessment indicated no challenges with audio-based evaluation. For instance, one worker stated that "all the audio samples were good," while another noted that "the audio is very clear". A general comment read, "the HIT is very unique, and there were no issues during the experiment". These preliminary results, obtained using non-expert crowd workers, suggest the effectiveness of speech in MT evaluation. However, further investigation may be required, and a more fine-grained approach—such as error annotation—could help better quantify the impact of audio in MT assessment.

\section{Conclusion}

We have presented our findings on integrating speech into human evaluation of MT quality. Our experiments with crowd workers compared MT system rankings from text-only and speech-enabled evaluation setups. 

Despite using basic TTS tools and crowd workers, our study extends MT evaluation beyond traditional text-based assessments, highlighting the potential of audio-based evaluation to provide distinct insights into MT evaluation. As MT research increasingly embraces multimodal translation, our findings provide empirical evidence that text-only evaluation may be insufficient. Beyond MT, this approach could benefit fields such as automatic dubbing, AI-assisted interpreting, and multilingual speech interfaces. Overall, our study emphasizes the need for more holistic evaluation benchmarks that better reflect the complexity of real-world language use.

Our code\footnote{\url{https://github.com/sami-haq99/Multimodal_Direct_Assessment}} and collected human annotation data are freely available.




\section{Limitations}
We performed general adequacy-based MT evaluation using crowd-workers on a limited dataset. Since the primary goal was to test whether audio-based judgments make a difference, we employed a simplified assessment approach and focused only on the English–German language pair.
We acknowledge that even expert-based human judgments can be noisy, potentially leading to low inter-annotator agreement (IAA) if not carefully conducted. Nevertheless, we collected a large sample of annotations from crowd-workers to compare the two approaches. With intrinsic quality control measures, crowd-sourced annotations have been shown to achieve higher IAA \citep{graham_can_2017}. However, due to limited time and platform constraints, manual filtering of noisy annotations was not feasible, making it difficult to eliminate all low-quality responses.

As MT quality evaluation has increasingly moved toward ESA-style \cite{kocmi2024error} annotations (at least in WMT), audio-based evaluation could be integrated into such platforms to identify error spans by listening to translations and assigning final scores. However, accurately segmenting the audio for this purpose would pose a significant challenge. 

\section{Ethical Considerations}

The human annotations collected via Amazon Mechanical Turk were fully anonymous. Anonymous users with MTurk accounts (meeting the defined criteria) submitted the tasks using numeric worker IDs. Although no personal identity information was revealed, we removed the worker IDs after payments were processed. Since the crowd-workers only needed to be native speakers and were not required to be expert translators, they were compensated according to the platform’s minimum task rate.

In cases where crowd-workers did not meet the quality control criteria—such as submitting robotic responses or completing tasks in an unrealistically short time—we rejected their submissions and did not provide payment. 

\section*{Acknowledgements}
This work was conducted with the financial support of the Research Ireland Centre for Research Training in Digitally-Enhanced Reality (d-real) under Grant No. 18/CRT/6224, and Research Ireland Centre for Research Training in Artificial Intelligence under Grant No. 18/CRT/6223. For the purpose of Open Access, the author has applied a CC BY public copyright licence to any Author Accepted Manuscript version arising from this submission.

The Authors also benefit from being members of the ADAPT SFI Research Centre at Dublin City University, funded by the Science Foundation Ireland under Grant Agreement No. 13/RC/2106\_P2.

\bibliography{emnlp2023-latex/custom}
\bibliographystyle{acl_natbib}

\clearpage 
\onecolumn
\appendix
\renewcommand{\thesection}{\Alph{section}}
\section{Head to Head Significance test Results}
\label{appendix_head}
The following tables (\ref{head-text}--\ref{head_official}) show differences in average standardized human scores for a system in that column and the system in that row for the German--English language pair. We applied the Wilcoxon rank-sum test to measure the likelihood that such differences could occur simply by chance for text-only, text-audio and WTM22 official\footnote{For official results, we used the human evaluation data provided by WMT22 organisers at: \url{https://github.com/wmt-conference/wmt22-news-systems/tree/main/humaneval/DA}. } experiments. In the following tables, * indicates statistical significance at $p < 0.05$, ** indicates statistical significance at $p < 0.01$, and *** indicates statistical significance at $p < 0.001$, according to the Wilcoxon rank-sum test. 

Each table contains a final column showing the total number of judgments used to calculate the results. The number for the official results is much greater than in our experiments; therefore, a direct comparison should only be made between the text-only and multimodal scores.

\begin{table}[H]
\centering
\small 
\setlength{\tabcolsep}{5pt} 
\renewcommand{\arraystretch}{1.5} 
\begin{tabular}{r|*{10}{c}|c}

 & \rotatebox{90}{HUMAN} & \rotatebox{90}{JDExploreAcademy} & \rotatebox{90}{LT22} & \rotatebox{90}{Lan-Bridge} & \rotatebox{90}{Online-A} & \rotatebox{90}{Online-B} & \rotatebox{90}{Online-G} & \rotatebox{90}{Online-W} & \rotatebox{90}{Online-Y} & \rotatebox{90}{PROMT} & \rotatebox{90}{No. of Judgments} \\
\hline
HUMAN & -- & -0.19 & -0.20 & -0.20 & -0.31 & -0.28 & -0.19 & -0.38 & -0.27 & -0.31 & 371 \\
JDExploreAcademy & 0.19** & -- & -0.01 & -0.01 & -0.13 & -0.09 & 0 & -0.19 & -0.08 & -0.12 & 385 \\
LT22 & 0.20*** & 0.01 & -- & 0 & -0.12 & -0.08 & 0.01 & -0.18 & -0.07 & -0.11 & 475 \\
Lan-Bridge & 0.20** & 0.01 & 0 & -- & -0.12 & -0.09 & 0.01 & -0.18 & -0.07 & -0.11 & 399 \\
Online-A & 0.31*** & 0.13** & 0.12* & 0.12* & -- & 0.03 & 0.13** & -0.07 & 0.04 & 0.01 & 451 \\
Online-B & 0.28*** & 0.09 & 0.08 & 0.09 & -0.03 & -- & 0.09 & -0.10 & 0.01 & -0.02 & 427 \\
Online-G & 0.19** & 0 & -0.01 & -0.01 & -0.13 & -0.09 & -- & -0.19 & -0.08 & -0.12 & 385 \\
Online-W & 0.38*** & 0.19** & 0.18* & 0.18** & 0.07 & 0.10* & 0.19*** & -- & 0.11* & 0.08 & 433 \\
Online-Y & 0.27*** & 0.08 & 0.07 & 0.07 & -0.04 & -0.01 & 0.08 & -0.11 & -- & -0.03 & 417 \\
PROMT & 0.31*** & 0.12* & 0.11 & 0.11 & -0.01 & 0.02 & 0.12** & -0.08 & 0.03 & -- & 349 \\

\end{tabular}
\caption{Head to Head comparison matrix of text-only judgments with significance levels and number of judgments.}
\label{head-text}
\end{table}

\begin{table}[H]
\centering
\small
\setlength{\tabcolsep}{5pt} 
\renewcommand{\arraystretch}{1.5} 
\begin{tabular}{r|*{10}{c}|c}

 & \rotatebox{90}{HUMAN} & \rotatebox{90}{JDExploreAcademy} & \rotatebox{90}{LT22} & \rotatebox{90}{Lan-Bridge} & \rotatebox{90}{Online-A} & \rotatebox{90}{Online-B} & \rotatebox{90}{Online-G} & \rotatebox{90}{Online-W} & \rotatebox{90}{Online-Y} & \rotatebox{90}{PROMT} & \rotatebox{90}{No. of Judgments} \\
\hline
HUMAN & -- & -0.24 & -0.26 & -0.21 & -0.36 & -0.24 & -0.39 & -0.33 & -0.32 & -0.35 & 445 \\
JDExploreAcademy & 0.24*** & -- & -0.02 & 0.04 & -0.12 & 0 & -0.15 & -0.09 & -0.07 & -0.11 & 426 \\
LT22 & 0.26*** & 0.02 & -- & 0.06 & -0.10 & 0.02 & -0.13 & -0.07 & -0.05 & -0.09 & 470 \\
Lan-Bridge & 0.21** & -0.04 & -0.06 & -- & -0.15 & -0.04 & -0.18 & -0.13 & -0.11 & -0.15 & 514 \\
Online-A & 0.36*** & 0.12* & 0.10* & 0.15** & -- & 0.12* & -0.03 & 0.02 & 0.04 & 0 & 435 \\
Online-B & 0.24*** & 0 & -0.02 & 0.04 & -0.12 & -- & -0.14 & -0.09 & -0.07 & -0.11 & 499 \\
Online-G & 0.39*** & 0.15* & 0.13* & 0.18*** & 0.03 & 0.14* & -- & 0.05 & 0.07 & 0.03 & 487 \\
Online-W & 0.33*** & 0.09 & 0.07 & 0.13* & -0.02 & 0.09 & -0.05 & -- & 0.02 & -0.02 & 464 \\
Online-Y & 0.32*** & 0.07 & 0.05 & 0.11* & -0.04 & 0.07 & -0.07 & -0.02 & -- & -0.04 & 394 \\
PROMT & 0.35*** & 0.11* & 0.09* & 0.15** & 0 & 0.11* & -0.03 & 0.02 & 0.04 & -- & 549 \\

\end{tabular}
\caption{Head-to-head comparison matrix of multimodal annotations with significance levels and number of judgments.}
\label{head_multimodal}
\end{table}

\FloatBarrier
\begin{table}[H]
\centering
\small
\setlength{\tabcolsep}{5pt} 
\renewcommand{\arraystretch}{1.5} 
\begin{tabular}{r|*{10}{c}|c}

 & \rotatebox{90}{HUMAN-B} & \rotatebox{90}{JDExploreAcademy} & \rotatebox{90}{LT22} & \rotatebox{90}{Lan-Bridge} & \rotatebox{90}{Online-A} & \rotatebox{90}{Online-B} & \rotatebox{90}{Online-G} & \rotatebox{90}{Online-W} & \rotatebox{90}{Online-Y} & \rotatebox{90}{PROMT} & \rotatebox{90}{No. of Judgments} \\
\hline
HUMAN-B & -- & -0.04 & 0.05*** & -0.07 & -0.01 & 0.01 & -0.02 & -0.05 & 0.03* & 0.07** & 2100 \\
JDExploreAcademy & 0.04 & -- & 0.09*** & -0.03 & 0.03 & 0.05* & 0.02 & -0.01 & 0.07** & 0.11*** & 2100 \\
LT22 & -0.05 & -0.09 & -- & -0.12 & -0.06 & -0.04 & -0.07 & -0.10 & -0.03 & 0.02 & 2100 \\
Lan-Bridge & 0.07** & 0.03* & 0.12*** & -- & 0.06*** & 0.08*** & 0.05** & 0.02* & 0.09*** & 0.14*** & 2100 \\
Online-A & 0.01 & -0.03 & 0.06** & -0.06 & -- & 0.02 & -0.01 & -0.04 & 0.04 & 0.08** & 2100 \\
Online-B & -0.01 & -0.05 & 0.04* & -0.08 & -0.02 & -- & -0.03 & -0.06 & 0.02 & 0.06* & 2100 \\
Online-G & 0.02 & -0.02 & 0.07*** & -0.05 & 0.01 & 0.03 & -- & -0.03 & 0.04* & 0.09** & 2100 \\
Online-W & 0.05 & 0.01 & 0.10*** & -0.02 & 0.04 & 0.06* & 0.03 & -- & 0.07*** & 0.12*** & 2100 \\
Online-Y & -0.03 & -0.07 & 0.03 & -0.09 & -0.04 & -0.02 & -0.04 & -0.07 & -- & 0.05 & 2100 \\
PROMT & -0.07 & -0.11 & -0.02 & -0.14 & -0.08 & -0.06 & -0.09 & -0.12 & -0.05 & -- & 2100 \\

\end{tabular}
\caption{Head-to-head comparison matrix of WMT22 official rankings with significance levels and number of judgments.}
\label{head_official}
\end{table}

\end{document}